\documentclass[letterpaper]{article} % DO NOT CHANGE THIS
\usepackage[draft]{aaai2027}  % DO NOT CHANGE THIS
\usepackage[hyphens]{url}  % DO NOT CHANGE THIS
\usepackage{graphicx} % DO NOT CHANGE THIS
\usepackage{natbib}  % DO NOT CHANGE THIS AND DO NOT ADD ANY OPTIONS TO IT
\usepackage{caption} % DO NOT CHANGE THIS AND DO NOT ADD ANY OPTIONS TO IT
\usepackage{booktabs}

\usepackage{amsfonts}
\usepackage{amsmath}
\usepackage{nicefrac}
\usepackage{microtype}
\usepackage[table]{xcolor}
\usepackage{multirow}
\usepackage{adjustbox}
\usepackage{makecell}
\usepackage{placeins}
\usepackage{tikz}
\usetikzlibrary{positioning,arrows.meta,shapes.geometric,calc,fit,backgrounds}

\definecolor{bestred}{RGB}{204,0,0}
\definecolor{secondblue}{RGB}{0,0,204}
\definecolor{oursrow}{gray}{0.93}

\newcommand{\best}[1]{\textcolor{bestred}{\textbf{#1}}}
\newcommand{\second}[1]{\textcolor{secondblue}{\underline{#1}}}
\newcommand{\ours}[1]{\textbf{#1}}  % our method emphasis
\newcommand{\keywords}[1]{\par\noindent\small\vphantom{\textbf{Keywords:} #1}\par\vspace{0.5ex}}

\title{PCQA-R1: Advancing Generalized 3D Point Cloud Quality Assessment with Reinforcement Learning}

\author{
    Kangning Ye\textsuperscript{\rm 1}\equalcontrib,
    Yunhao Li\textsuperscript{\rm 1}\equalcontrib,
    Sijing Wu\textsuperscript{\rm 1},
    Yucheng Zhu\textsuperscript{\rm 1},
    Guangtao Zhai\textsuperscript{\rm 1}
}
\affiliations{
    \textsuperscript{\rm 1}Shanghai Jiao Tong University
}

\begin{document}
\raggedbottom 
\maketitle

\begin{abstract}
No-reference point cloud quality assessment (PCQA) has been an active topic in recent years and is used to measure and optimize the visual experience of point clouds. However, large multimodal models (LMMs) have rarely been explored in this area. Previous LMM-based methods mainly rely on supervised fine-tuning to directly predict numerical quality scores, lacking the ability to generalize across datasets with heterogeneous MOS scales and limited annotations. A key difficulty is that absolute MOS regression can be brittle across datasets with different score scales and distortion distributions, whereas relative quality ranking is more stable under such shifts. In this paper, we present \textbf{PCQA-R1}, the first reinforcement learning LMM for 3D point cloud quality assessment to simultaneously model quality understanding and scoring. Built upon the group relative policy optimization (GRPO) strategy, PCQA-R1 first constructs a chain-of-thought dataset, PCQA-CoT, which serves as cold-start training data through a reverse reasoning strategy that teaches the LMM to generate its reasoning process. We further introduce a Gaussian proximity reward that prevents calibration drift by anchoring score predictions to the source MOS range. Experimental results demonstrate that PCQA-R1 achieves state-of-the-art cross-dataset generalization across five benchmarks and competitive in-domain accuracy. Ablation studies support the role of ranking, Gaussian reward, and cold-start traces.

\end{abstract}

% \keywords{\hspace{0pt}} 

\begin{figure*}[!t]
\centering
\includegraphics[width=\textwidth]{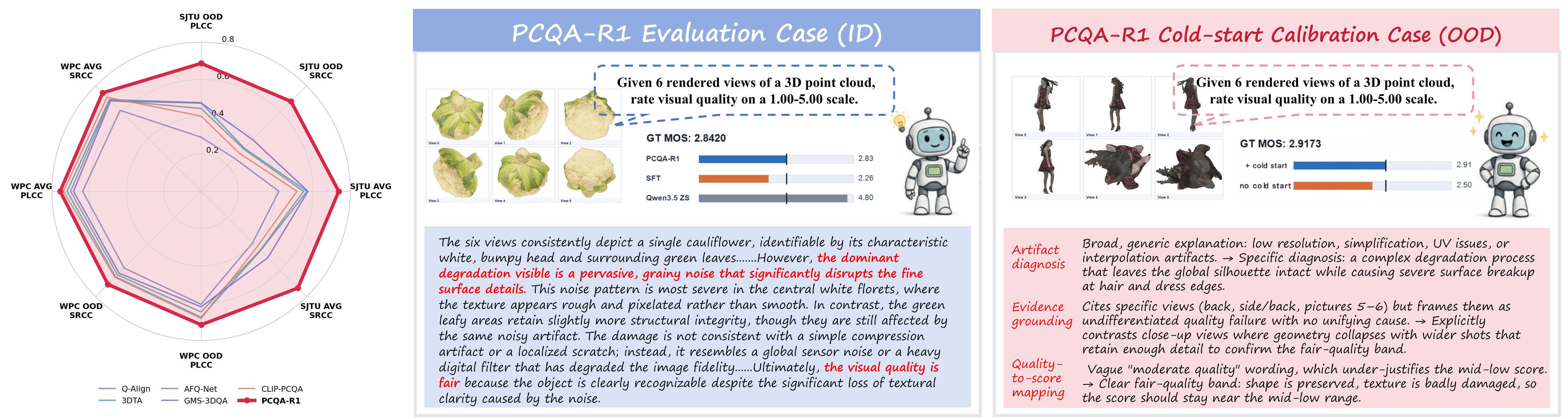}
\caption{%
  \textbf{Cross-domain trend and qualitative evidence.}
  (Left) Radar plot of PLCC/SRCC transfer from SJTU and WPC sources. (Right) PCQA-R1 mapping from six-view cues to quality score. The case study is qualitative only; fuller excerpts are in Appendix~E in the supplementary material.
}
\label{fig:teaser}
\end{figure*}

% ===========================================================================
\section{Introduction}
\label{sec:intro}

Three-dimensional (3D) point clouds have become a dominant representation for
immersive media, autonomous driving, and digital twins, where end-to-end
processing pipelines (acquisition, compression, transmission, rendering)
inevitably degrade perceived quality.  Objective \emph{point cloud quality
assessment} (PCQA) is therefore crucial for calibrating codecs, guiding
streaming decisions, and benchmarking 3D reconstructions.  The core challenge we address is cross-dataset generalization: projection
models trained on SJTU-PCQA achieve up to $0.97$~PLCC in-domain but drop to
as low as $0.28$~SRCC on LS-PCQA, because they learn dataset-specific MOS
scales rather than transferable quality relations. This paper asks: can reinforcement learning learn quality rankings that transfer across datasets without seeing target-domain data?

Existing PCQA methods have advanced along several lines: full-reference
geometry/color metrics, projection-based no-reference (NR) models, 3D-native
or multimodal hybrids, and recent large multimodal model (LMM) scorers.  Strong
projection models such as CLIP-PCQA~\cite{liu2025clip}, GMS-3DQA~\cite{zhang2024gms},
and AFQ-Net~\cite{zhang2024asynchronous} achieve impressive in-domain
correlations, while LMM methods such as Q-Align~\cite{wu2023q} make large
multimodal backbones usable for PCQA.
However, most of these methods still learn by regressing or classifying
dataset-specific mean opinion scores (MOS).  They can output a number, but they
usually do not expose which visible defects, viewpoints, or structure changes
justify that number.  This lack of explicit quality evidence is not only a
trust issue: under cross-dataset evaluation, the same regression target must
absorb different MOS scales, distortion taxonomies, object distributions, and
projection conventions, so in-domain accuracy often fails to transfer
(Tables~\ref{tab:cross_sjtu},~\ref{tab:cross_wpc}).

We address this gap with two design choices.  \textit{First}, we study
structured quality evidence as part of the learning problem, rather than as
post-hoc decoration.  Recent image-quality work such as
Q-Insight~\cite{li2026q} argues that a quality model should connect visible
artifacts, content preservation, degradation severity, and the final rating in
its reasoning process.  For PCQA, this suggests a concrete hypothesis:
view-specific degradation cues --- compression artifacts, color corruption,
geometry breakage, missing structures, and surface defects --- are more
transferable than an absolute MOS anchor.  Point clouds make this hypothesis
especially relevant because quality is inherently multi-view: a defect may be
obvious in one canonical view and nearly invisible in another.  A model that is
initialized to organize such evidence before scoring may rely less on
source-dataset score shortcuts.

Second, perceived quality is fundamentally \emph{relative}: a ranking signal is
naturally invariant to monotone MOS rescaling across datasets, motivating a
Thurstone-style GRPO reward~\cite{guo2025deepseek,wu2026visualquality}.
We ask: \emph{can an LMM learn transferable quality rankings from source-domain
data alone?}  Concurrent QD-PCQA~\cite{zhang2026qd} targets the same goal via
domain adaptation with unlabeled target data; PCQA-R1 uses no target samples.

We answer this question with \textbf{PCQA-R1}, a reinforcement-learning-to-rank
framework for projection-based PCQA.  Each point cloud is represented by six
canonical color views and a single scoring prompt.  The model does not receive
depth, normals, point patches, or explicit 3D feature embeddings; all reported
results use the same color-view input as the LMM baselines.  PCQA-R1 fine-tunes
Qwen3.5-9B~\cite{team2026qwen3} with GRPO~\cite{guo2025deepseek} using a ranking-primary
reward bundle.  The main training recipe adds a Gaussian proximity reward for
source-scale score shaping and a short SFT cold start with sample-specific
quality-analysis traces.  These traces ask the policy to
organize evidence from the six rendered views --- object structure, geometry
preservation, color fidelity, compression artifacts, and local defects ---
before emitting a score, so RL begins from an artifact-aware scoring policy
rather than from a pure score regressor.  The final answer is still a precise
score, but the route to that score is optimized by verifiable ranking and
score-shaping rewards.
Unlike VisualQuality-R1~\cite{wu2026visualquality}, which trains on
$14{,}000{+}$ 2D images where ranking alone stabilises, PCQA-R1 faces a
$30$--$40{\times}$ smaller MOS-labelled training set and must aggregate
evidence across six views of irregular 3D geometry---motivating both the
Gaussian proximity reward and the cold-start initialisation that are
unnecessary at 2D scale.

\noindent\textbf{Contributions.}
\begin{itemize}
\item \textbf{A ranking-based LMM-RL adaptation for PCQA.}  We formulate NR-PCQA
as reinforcement learning to rank, using a Qwen3.5-9B LMM to score six colour
views without depth maps, point patches, or explicit 3D feature injection.
To our knowledge, this is the first reinforcement-learning LMM formulation for NR-PCQA. A rank-invariance property of the Thurstone reward provides a theoretical basis for its cross-domain stability.
\item \textbf{A score-shaping reward and cold-start recipe.}  Thurstone-style GRPO,
Gaussian proximity, and a one-epoch quality-trace warm-up jointly separate rank
transfer from source-scale score alignment.
\item \textbf{Empirical evidence for ranking-based generalization.}  Fold-controlled
experiments across five benchmarks and two source domains show stronger average
OOD transfer, with ablations isolating ranking, calibration, and cold start.
\end{itemize}

% ===========================================================================
\section{Related Work}
\label{sec:related}
% ===========================================================================

\subsection{Reinforcement Learning for Vision--Language Reasoning}
\label{sec:related_rl_vqa}
DeepSeek-R1~\cite{guo2025deepseek} popularized GRPO, which normalizes sampled
responses within each prompt group and removes the need for a critic.  Vision
extensions use similarly verifiable rewards for grounding~\cite{shen2025vlm},
image quality~\cite{li2026q,wu2026visualquality,cai2025unified,feng2025preresq,cao2026qualirag,li2025exploring,wu2026q,li2025aghi},
and video quality~\cite{cao2026vqathinker,li2026videoaesbench,wu2025fvq,wu2025hveval}.  These works show that reasoning
traces help when tied to task feedback, but they remain 2D image/video tasks.
Recent survey work further documents how large-model-based quality assessment
is shifting from direct score prediction toward more reasoning-aware pipelines~\cite{zhang2025quality}.
VisualQuality-R1~\cite{wu2026visualquality}
trains on 14K+ 2D image samples with a Thurstone ranking reward.
PCQA-R1 operates on 336--592 labelled samples per domain---one to two orders
of magnitude smaller---and must aggregate evidence across six rendered views of
irregular 3D geometry.  To stabilize training at this scale, we pair ranking
with a Gaussian proximity reward and a 1-epoch cold-start trace initialization,
neither of which appears necessary in the larger-data 2D setting.

\begin{figure*}[!t]
  \centering
  \includegraphics[width=\linewidth]{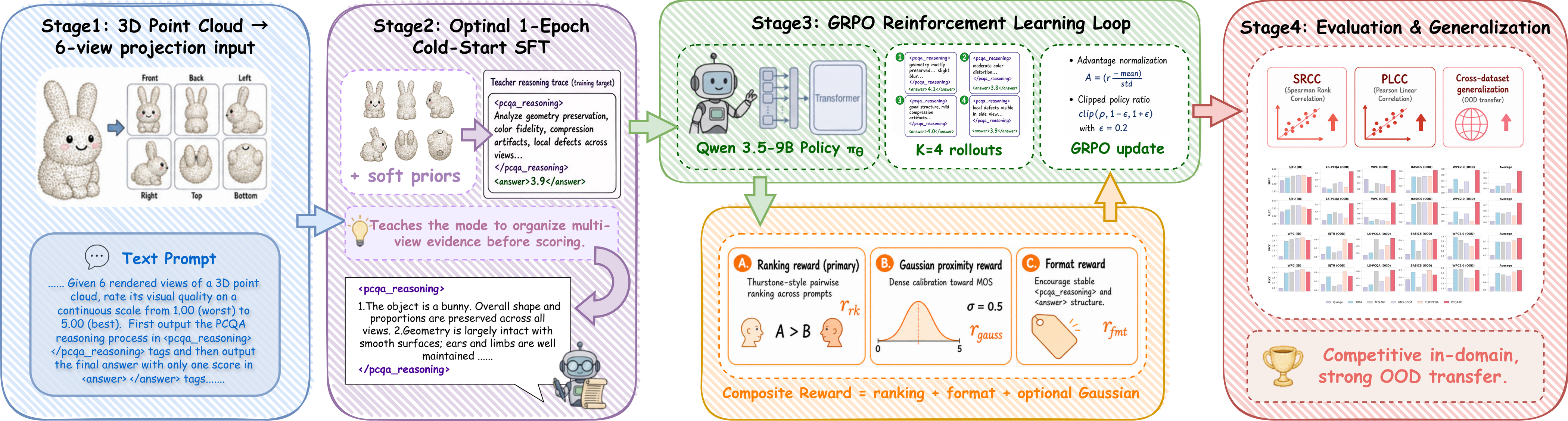}
  \caption{%
    \textbf{Overview of PCQA-R1.}
    Six canonical colour views and a scoring prompt are fed to a Qwen3.5-9B
    policy.  For each prompt, $K{=}4$ rollouts produce a reasoning block and a
    score.  GRPO optimises ranking, an optional Gaussian score-shaping term, and format
    stability under KL regularisation; an optional one-epoch SFT cold start
    provides the initial quality-analysis traces.%
  }
  \label{fig:pipeline}
\end{figure*}

\subsection{3D Point Cloud Quality Assessment}
\label{sec:related_pcqa}
PCQA methods mainly differ in their input evidence.  \textbf{FR metrics}
compare a distorted cloud with its reference using geometry or appearance cues,
including PCQM~\cite{meynet2020pcqm}, GraphSIM~\cite{yang2020inferring},
PointSSIM~\cite{alexiou2018point}, PointPCA~\cite{alexiou2024pointpca}, PSNR$_{\text{YUV}}$, and recent MLLM-assisted
FR scoring~\cite{watanabe2025full}.  They are informative when a
pristine cloud exists but are unusable for reference-free capture or streaming.
\textbf{NR projection methods} reuse 2D backbones over rendered views, from
PQA-Net, 3DTA, GMS-3DQA, AFQ-Net, and CLIP-PCQA~\cite{liu2021pqa, zhu20243dta, zhang2024gms, wu2026fmreward, zhang2024asynchronous, liu2025clip} to saliency-guided and transfer/alignment variants~\cite{zhou2024visual,su2025progressive,wu2025net}. They are strong in-domain but often inherit dataset-specific MOS scales. \textbf{3D
or multimodal models} add sparse 3D CNNs, point patches, cross-media transfer,
cross-modal contrastive enhancement~\cite{ming2025no}, or dynamic video evidence~\cite{liu2023point,zhang2022mm,zhang2022no,liu2025images,yang2025mpv,chakraborty2025mt},
yet still learn by MOS regression.  Related work on dynamic 4D human quality~\cite{li2026dhqa,wu2025singinghead}
and multi-dimensional video quality~\cite{wu2026multi} similarly employs MOS supervision but for different 3D/4D modalities.
\textbf{LMM-based PCQA} uses multimodal
foundation models for scoring or graph-assisted reasoning~\cite{wu2023q,zhang2024lmm,gupta2025pit,xie2024llm,duan2026bmpcqa},
but remains supervised by absolute scores.

The closest cross-domain line is domain adaptation, including early PCQA
adaptation~\cite{yang2022no}, UPDA~\cite{xie2026upda}, and
QD-PCQA~\cite{zhang2026qd}.  These methods align source and target distributions,
often requiring unlabeled target samples.  PCQA-R1 instead changes the learning
signal itself: it optimizes relative ranking with no target-domain data.
The broader shift toward reasoning-aware LMM quality assessment~\cite{li2026q,cai2025unified,feng2025preresq,cao2026qualirag,wu2026visualquality,zhang2025quality} motivates applying this to 3D point clouds.

% ===========================================================================
\section{Method}
\label{sec:method}
% ===========================================================================
Figure~\ref{fig:pipeline} summarizes PCQA-R1.  The method has three parts: a
fixed six-view input protocol, a GRPO objective centered on relative quality
ordering with an optional score-shaping term, and a short quality-trace cold start. We begin with GRPO preliminaries, then formalize the prediction task, and describe the rendering protocol, ranking reward, Gaussian score-shaping term, and cold-start stage.

\subsection{Preliminaries: Group Relative Policy Optimization}
\label{sec:prelim}
Group Relative Policy Optimization (GRPO)~\cite{guo2025deepseek} fine-tunes
large language or multimodal models with reward signals without requiring a
separate critic network.  For each training prompt $x$, GRPO samples $K$
independent completions $\{o^{(k)}\}_{k=1}^{K}$, evaluates scalar rewards
$\{r^{(k)}\}$, and computes within-group normalized advantages:
\begin{equation}
  \hat{A}^{(k)} = \frac{r^{(k)} - \mathrm{mean}_k(r)}
                       {\mathrm{std}_k(r) + \epsilon},
  \label{eq:grpo_adv}
\end{equation}
where $\epsilon$ is a small stability constant.  The policy is then updated
using a clipped-ratio objective with a KL penalty against a frozen reference
model $\pi_{\mathrm{ref}}$:
\begin{equation}
\begin{split}
  \mathcal{J}(\theta)
  &= \mathbb{E}_{k}\!\Bigl[
    \min\!\bigl(\rho^{(k)}\hat{A}^{(k)},\;
      \mathrm{clip}(\rho^{(k)},1{-}\varepsilon,1{+}\varepsilon)\hat{A}^{(k)}\bigr)
    \\
    &\quad\; - \beta\,D_{\mathrm{KL}}\bigl[\pi_\theta(\cdot|x)\,\|\,\pi_{\mathrm{ref}}(\cdot|x)\bigr]
  \Bigr],
\end{split}
  \label{eq:grpo_base}
\end{equation}
where $\rho^{(k)}\!=\!\pi_\theta(o^{(k)}|x)/\pi_{\theta_{\mathrm{old}}}(o^{(k)}|x)$
and $\varepsilon{=}0.2$.  The KL term prevents the policy drifting far from the
pre-trained model, regularising against reward hacking.  In PCQA-R1, the scalar
reward $r^{(k)}$ is replaced by a composite ranking reward that measures
batch-level quality ordering and score calibration; the full derivation is given in the GRPO reward subsection.

\subsection{Problem Formulation}
\label{sec:method_formulation}
Given a point cloud $\mathcal{P}$ with MOS $y \in [y_{\min}, y_{\max}]$, we
render it into an ordered tuple of six canonical colour views
$\mathbf{I}(\mathcal{P})=\{I_1,\dots,I_6\}$ and define a policy
$\pi_\theta(o\mid\mathbf{I},c)$ over a language output $o$ given a textual
prompt $c$.  The output follows the template
\texttt{<pcqa\_reasoning>$\cdot$</pcqa\_reasoning>}\allowbreak\texttt{<answer>$\hat{y}$</answer>}, from which a
scalar prediction $\hat{y}\in[y_{\min},y_{\max}]$ is parsed.  The training
objective is to maximise a multi-term reward that rewards \emph{ranking
consistency} across different point clouds within the same mini-batch, with
the expectation that a ranking-correct policy generalises across datasets.

\subsection{Multi-View Rendering Protocol}
\label{sec:method_views}
We adopt six canonical viewpoints (front, back, left, right, top, bottom)
rendered under one shared colour-view protocol applied identically across SJTU-PCQA~\cite{yang2020predicting},
LS-PCQA~\cite{liu2023point}, WPC~\cite{liu2022perceptual}, WPC2.0~\cite{liu2021reduced}, and BASICS~\cite{ak2024basics}.
Only colour views are used.  We do not inject explicit point features, depth,
normal maps, or mesh information in the reported experiments.  This
single-protocol choice allows direct checkpoint transfer between source and
target datasets; implementation details are given in Appendix~A of the supplementary material.

\begin{figure*}[!t]
  \centering
  \includegraphics[width=\linewidth]{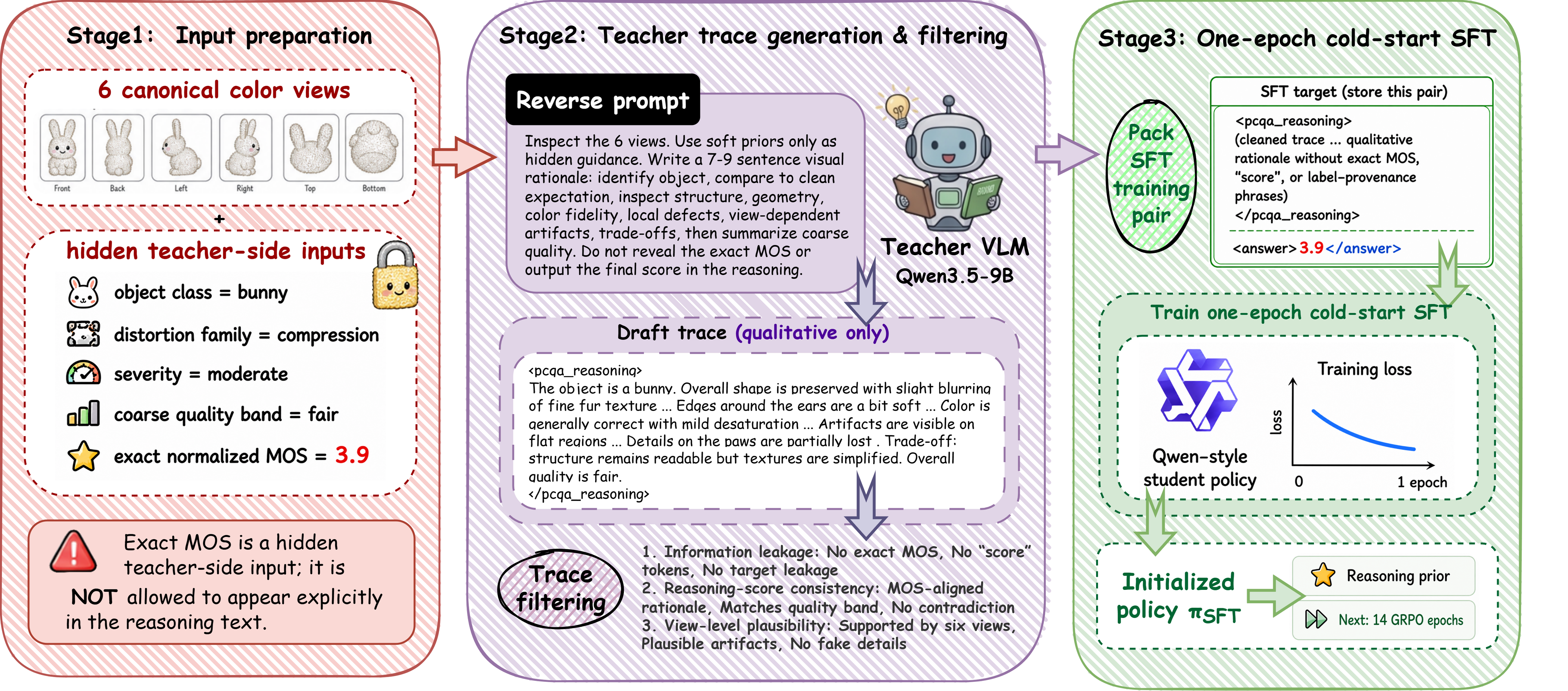}
  \caption{%
    \textbf{Cold-start trace construction and training use.}
    For source-training samples only, a Qwen3.5-9B teacher model receives the
    six rendered views, metadata-derived weak priors, and the exact normalized
    MOS as a hidden calibration signal.  The generated compact 7--9 sentence rationale
    is filtered for leakage and factual plausibility, then stored as the
    \texttt{<pcqa\_reasoning>} target, while MOS remains only in the final
    \texttt{<answer>} field during student SFT.  We treat these traces as
    score-calibrated training rationales, not as independent human
    explanations; the one-epoch warm start initializes artifact-aware quality
    reasoning before 14 epochs of GRPO ranking refinement.
  }
  \label{fig:coldstart}
\end{figure*}

\subsection{GRPO with Thurstone Ranking Reward}
\label{sec:method_grpo}
For each prompt $x_i$ (a point cloud plus its 6-view stack), the policy
samples $K=4$ completions $\{o^{(k)}_i\}_{k=1}^K$, from which scores
$\{\hat{y}^{(k)}_i\}$ are parsed.  Let $\mu_i,\sigma_i$ denote the per-prompt
empirical mean and standard deviation of $\{\hat{y}^{(k)}_i\}$.  GRPO needs a
separate reward for every sampled completion, not only one reward for the
prompt mean.  We therefore use a rollout-conditioned Thurstone-style
comparison: rollout $k$ of prompt $i$ is represented by
$Q_i^{(k)}\!\sim\!\mathcal{N}(\hat{y}^{(k)}_i,\sigma_i^2)$, while another
prompt $j$ is summarized by its rollout consensus
$Q_j\!\sim\!\mathcal{N}(\mu_j,\sigma_j^2)$.  Under independent Gaussian
noise, the probability that rollout $k$ of prompt $i$ is rated higher than
prompt $j$ is
\begin{equation}
  p_k(x_i,x_j) \;=\;
  \Phi\!\left(
    \frac{\hat{y}^{(k)}_i - \mu_j}{\sqrt{\sigma_i^2+\sigma_j^2+\gamma}}
  \right),
  \label{eq:thurstone}
\end{equation}
where $\Phi(\cdot)$ is the standard-normal CDF and $\gamma{=}10^{-6}$.
The asymmetric numerator keeps completions from the same prompt separable:
different rollouts $\hat{y}^{(k)}_i$ can receive different ranking rewards,
while $\mu_j$ provides a stable batch-level anchor for the comparison prompt.
For tractability, all rollouts from the same prompt share the same variance
$\sigma_i^2$, estimated from the $K$ sampled scores; our goal here is to model
prompt-level dispersion rather than rollout-specific heteroscedastic noise.
This is not a prompt-mean comparison $P(Q_i>Q_j)$ using $\mu_i$ against
$\mu_j$; it is the rollout-level form used by VisualQuality-R1 and by our
training implementation (Appendix~A).
Eq.~\ref{eq:thurstone} should therefore be read as an asymmetric
rollout-conditioned approximation rather than the classical symmetric
mean-vs-mean Thurstone model.

Denoting the GT ranking probability
$g_{ij}=\mathbb{1}[y_i{>}y_j]$, the fidelity ranking reward is
\begin{equation}
  r^{(k)}_{\text{rk}}(x_i) \;=\;
  \frac{1}{|\mathcal{Z}_i|}\!\!\sum_{j\in\mathcal{Z}_i}\!
  \Bigl[\sqrt{p_k\,g_{ij}}+\sqrt{(1{-}p_k)(1{-}g_{ij})}\Bigr],
  \label{eq:fidelity}
\end{equation}
where $\mathcal{Z}_i$ is the set of other unique prompts in the batch.
Valid gradients require $|\mathcal{Z}_i|{\ge}1$, which translates to the
practical constraint that the per-device batch satisfies
$\text{batch/GPU}\ge 2K$.

\paragraph{Rank-invariance property.}
The fidelity reward $r^{(k)}_{\text{rk}}$ depends on ground-truth labels solely
through the ordinal indicator $g_{ij}=\mathbb{1}[y_i{>}y_j]$.  Because any
strictly increasing function $f$ preserves strict inequalities, replacing each
MOS $y_i$ with $f(y_i)$ leaves every $g_{ij}$ unchanged and hence leaves
$r^{(k)}_{\text{rk}}$ unchanged.  In particular, differences in MOS range or
distributional shift between source and target datasets do not corrupt the
ranking signal, which explains why the ranking reward remains robust to
cross-dataset MOS scale shifts.

\paragraph{Format reward.}
We reward outputs that contain a well-structured
\texttt{<pcqa\_reasoning>...</pcqa\_reasoning>} block followed by
\texttt{<answer>...</answer>}.
The reasoning block must contain nontrivial alphabetic or CJK content.  The
format reward is $r_{\text{fmt}}{=}1$ when this structure is matched and $0$
otherwise.  In the implementation, this term is an independent bonus rather
than a hard gate.

\paragraph{Combined reward.}
\begin{equation}
  r^{(k)} \;=\;
  r^{(k)}_{\text{rk}} \; +\; w_{\text{fmt}}(t)\,r_{\text{fmt}} \; +\; \alpha\,r^{(k)}_{\text{gs}},
  \qquad \alpha\in\{0,1\},
  \label{eq:total_reward}
\end{equation}
where the Gaussian proximity reward
$r^{(k)}_{\text{gs}}$ is introduced next, $\alpha{=}0$ for the ranking-only
ablation, and the training script uses a simple format curriculum:
$w_{\text{fmt}}(t){=}1$
for the first 20\% of GRPO steps---ensuring stable output structure before
ranking dominates---and then decays linearly to $0.3$ by the end of training.

\paragraph{GRPO policy objective.}
Let $A^{(k)}_i{=}\bigl(r^{(k)}_i-\mu_r^{(i)}\bigr)/\sigma_r^{(i)}$ be the
per-prompt normalised advantage, where $\mu_r^{(i)},\sigma_r^{(i)}$ are the
mean and standard deviation of $\{r^{(k)}_i\}_{k=1}^K$.  Denoting the
likelihood ratio
$\rho^{(k)}_i{=}\pi_\theta(o^{(k)}_i|x_i)/\pi_{\theta_{\text{old}}}(o^{(k)}_i|x_i)$,
We use the following GRPO objective.
\begin{equation}
\label{eq:grpo}
\begin{aligned}
  \mathcal{J}_{\text{GRPO}}(\theta)
  \;=\;
  \mathbb{E}_{i,k}\!
  \Bigl[&\min\!\bigl(
    \rho^{(k)}_i A^{(k)}_i,\;
    \mathrm{clip}(\rho^{(k)}_i,\,1{-}\varepsilon,\,1{+}\varepsilon)\,A^{(k)}_i
  \bigr)
  \\[-2pt]
  &\quad-\;
  \beta\,\mathrm{KL}\!\bigl(\pi_\theta(\cdot|x_i)\,\|\,
                          \pi_{\text{ref}}(\cdot|x_i)\bigr)
  \Bigr],
\end{aligned}
\end{equation}
with $\varepsilon{=}0.2$ and $\beta{=}0.04$.  Note that
the advantage is baselined \emph{within} each prompt, so the fidelity
ranking reward of Eq.~\eqref{eq:fidelity} --- which compares each rollout to
\emph{other} prompts' means --- supplies the between-prompt signal that a
per-prompt baseline cannot.

\subsection{Gaussian Proximity Reward}
\label{sec:method_gaussian}
In a low-resource PCQA setting (336--592 training samples per source
domain), sparse pairwise ranking rewards cannot prevent calibration
drift: a model may learn to rank correctly within training batches
while systematically anchoring predictions to a shifted score scale.
We add a Gaussian proximity reward,
\begin{equation}
  r^{(k)}_{\text{gs}}(x_i) \;=\; s\cdot\exp\!\Bigl(-
  \tfrac{(\hat{y}^{(k)}_i-y_i)^2}{2\sigma^2}\Bigr),
  \label{eq:gaussian}
\end{equation}
with $\sigma{=}0.5$ and $s{=}1.0$ in all experiments (MOS is normalized
to $[0,1]$ for every dataset), to provide a weak absolute anchor that
keeps score predictions calibrated to the MOS range during RL exploration.  The term decays to zero far from
$y_i$, preventing it from dominating the ranking objective while
correcting cross-domain scale mismatch.

\subsection{Cold-Start SFT with Quality-Analysis Traces}
\label{sec:method_coldstart}
Starting directly from the base model leaves two things for sparse RL to learn
at once: the PCQA reasoning style and the score-ranking policy.  We therefore
optionally prepend a one-epoch SFT stage, used as a \textit{reasoning prior}
rather than as the final learner.  We call the resulting trace collection
\textbf{PCQA-CoT}.  For each MOS-labeled point cloud, we pair the sample with
a generated compact 7--9 sentence quality-analysis trace that
first identifies stable visual evidence across the six canonical views and then
connects geometry, colour, and texture degradations to a coarse quality band
before the final MOS answer.
This teaches the model to organize viewpoint-aware evidence inside
\texttt{<pcqa\_reasoning>} before producing the numeric score in
\texttt{<answer>}.

The trace generator follows a reverse-thinking protocol.
We use \textbf{Qwen3.5-9B} as the teacher model, sharing the same base
architecture as the student.  For source-training samples only, it observes
the six rendered views,
metadata-derived priors such as object class, distortion family, severity, and
coarse quality band, and the exact normalized ground-truth MOS.  The teacher
uses the exact MOS only as a hidden calibration signal during offline trace
synthesis; it is not used for target-domain evaluation.  The prompt explicitly
forbids the teacher from writing the exact MOS, any numeric score, the word
``score'', or label-provenance phrases in the generated rationale, and the
stored trace is filtered for leakage before being used as the
\texttt{<pcqa\_reasoning>} target.  During student SFT, the exact MOS appears
only in the final \texttt{<answer>} field, as in standard supervised score
training.  Each stored trace is kept as a compact 7--9 sentence evidence chain
that follows a fixed order: object cue plus clean-reference expectation,
global preservation, dominant degradation, degradation justification,
damaged regions, preserved regions, and a coarse-band summary, with one or two
extra support sentences when needed.  The goal is not
to solve PCQA by score-conditioned rationale
imitation: the cold start only initializes a policy that can describe
transferable visual evidence, while the subsequent 14-epoch GRPO stage
optimizes relative ranking and calibrated score behavior from reward feedback.
Concretely, a 15-epoch SFT-only baseline achieves OOD AVG
SRCC~$0.649$ on SJTU, whereas 1-epoch cold-start followed by 14-epoch
GRPO reaches OOD AVG SRCC~$0.678$ ($+0.029$; Table~\ref{tab:ablation}),
suggesting that RL refinement, rather than teacher-generated trace supervision
alone, is the primary driver of cross-domain generalization.
We keep this warm-up short because excessive score-trace SFT can
over-specialize the language policy and narrow RL exploration.  The total
budget therefore remains 15 epochs: 1 SFT epoch plus 14 GRPO epochs.

% ===========================================================================
\section{Experiments}
\label{sec:experiments}
% ===========================================================================

\subsection{Experimental Setup}
\label{sec:setup}

All methods use the public \textbf{fold-1 split} from prior PCQA evaluation~\cite{liu2025clip}.
ID columns report the source test split; cross-dataset columns evaluate the full
target set (fold-0, no source-train overlap), except the WPC-source
WPC2.0$^{\dagger}$ protocol in Appendix~B.  We evaluate
SJTU-PCQA, LS-PCQA, WPC, WPC2.0, and BASICS~\cite{yang2020predicting,liu2023point,liu2022perceptual,liu2021reduced,ak2024basics}
with one canonical six-colour-view rendering protocol.  \best{Red bold} and
\second{blue underline} mark the best and runner-up NR methods; FR rows are
reference-only.

\paragraph{Evaluation.}
We report PLCC/SRCC for every dataset with the same source-selected
checkpoint and parsing pipeline.  Across the component comparisons, only the
ablated training module changes; split, decoding, and evaluation protocol are
held fixed.  Further metric and compute details are deferred to
Appendix~B and Appendix~A.

% ===========================================================================
\subsection{Cross-Dataset Generalization from SJTU}
% ===========================================================================

Table~\ref{tab:cross_sjtu} reports PLCC/SRCC when all no-reference methods
are trained on the small SJTU-PCQA fold-1 split (336 samples) and evaluated
zero-shot on four OOD target datasets.  The main comparison includes reproduced
fold-controlled baselines and full-reference reference metrics.
Unless otherwise noted, the PCQA-R1 row in Tables~\ref{tab:cross_sjtu}
and~\ref{tab:cross_wpc} denotes the full RL+Gaussian+Cold Start variant.
The RL+Gaussian row without cold start is retained in Table~\ref{tab:ablation}
for a controlled component comparison.

\begin{table*}[!t]
\centering
\setlength{\tabcolsep}{3pt}
\begin{tabular}{ll ccccc c}
\toprule
\multirow{2}{*}{\textbf{Category}} &
\multirow{2}{*}{\textbf{Method}} &
\textbf{SJTU} &
\textbf{LS-PCQA} &
\textbf{WPC} &
\textbf{BASICS} &
\textbf{WPC2.0} &
\multirow{2}{*}{\textbf{AVG.}} \\
 & & (ID) & (OOD) & (OOD) & (OOD) & (OOD) & \\
\midrule

% ---------- FR (Full-Reference) ----------
\multirow{5}{*}{FR}
 & p2point (RMS)
   & .758/.708
   & .410/.283
   & .458/.452
   & .559/.557
   & .461/.426
   & .529/.485 \\
 & p2plane
   & .665/.602
   & .394/.272
   & .378/.326
   & .472/.474
   & .425/.408
   & .467/.416 \\
 & GraphSIM~\cite{yang2020inferring}
   & .781/.746
   & .289/.227
   & .474/.394
   & .432/.394
   & .201/.186
   & .435/.389 \\
 & PointSSIM~\cite{alexiou2018point}
   & .686/.635
   & .398/.162
   & .424/.417
   & .363/.403
   & .475/.457
   & .469/.415 \\
 & PSNRyuv
   & .660/.655
   & .518/.494
   & .557/.539
   & .512/.457
   & .411/.408
   & .532/.511 \\

\midrule

% ---------- NR ----------
\multirow{8}{*}{NR}
 & IT-PCQA~\cite{zhang2022no}
   & .846/.608
   & .255/.150
   & .089/.040
  & .138/.047
   & .085/.013
  & .282/.172 \\
 & PQA-Net~\cite{liu2021pqa}
   & .784/.790
   & .227/.147
   & .321/.112
   & .462/.157
  & .181/.118
  & .395/.265 \\
 & ResSCNN~\cite{liu2023point}
   & .933/.928
   & .311/.312
   & .277/.260
   & .285/.214
   & .121/.068
   & .385/.356 \\
 & 3DTA~\cite{zhu20243dta}
   & .941/.940
   & .367/.261
   & .451/.136
   & .612/.447
   & \second{.471}/\second{.464}
   & .568/.450 \\
 & AFQ-Net$^{a}$~\cite{zhang2024asynchronous}
   & \second{.968}/.953
   & \second{.387}/.341
   & .479/.398
   & \best{.659}/.425
   & .252/.122
   & .549/.448 \\
 & CLIP-PCQA~\cite{liu2025clip}
   & .959/.957
   & .311/.286
   & .438/.300
   & \second{.658}/.454
  & .206/.119
  & .514/.423 \\
 & GMS-3DQA~\cite{zhang2024gms}
   & \best{.969}/\best{.958}
   & .370/\second{.364}
   & .457/.366
   & .653/\second{.456}
   & .414/.379
  & \second{.573}/\second{.505} \\
 & MM-PCQA$^{b}$~\cite{zhang2022mm}
   & .964/.958
   & .350/.296
   & .420/.283
   & .621/.398
   & .199/.174
   & .511/.422 \\

  % ---------- NR ----------
 & Q-Align~\cite{wu2023q}
   & .926/.946
   & .279/.279
   & \second{.529}/\second{.519}
   & .195/.071
   & .157/.138
   & .417/.391 \\

\midrule

% ---------- LMM-RL (Ours) ----------
\rowcolor{oursrow}
\textbf{LMM-RL}
 & \ours{PCQA-R1 (Ours)}
  & .954/.953
  & \best{.541}/\best{.547}
  & \best{.820}/\best{.826}
  & \best{.673}/\best{.634}
  & \best{.704}/\best{.704}
  & \best{.739}/\best{.733} \\

\bottomrule
\end{tabular}
\caption{%
  Cross-dataset PLCC/SRCC from \textbf{SJTU-PCQA} fold-1.  SJTU is ID; all other
  columns are zero-shot OOD targets.  Notes and provenance are in Appendix~B.
}
\label{tab:cross_sjtu}
\end{table*}

% ===========================================================================
\subsection{Cross-Dataset Generalization from WPC}
% ===========================================================================

Table~\ref{tab:cross_wpc} swaps the source dataset to WPC fold-1 (592 training
samples), which is compression-dominated (67.6\% compression-related distortions) on industrial
objects.  This setting stresses cross-content rather than cross-distortion
generalization: SJTU contains MPEG references with similar codecs but
different content, while LS-PCQA and BASICS introduce both new distortion
types and new content distributions.

\begin{table*}[!t]
\centering
\setlength{\tabcolsep}{3pt}
\begin{tabular}{ll ccccc c}
\toprule
\multirow{2}{*}{\textbf{Category}} &
\multirow{2}{*}{\textbf{Method}} &
\textbf{WPC} &
\textbf{SJTU} &
\textbf{LS-PCQA} &
\textbf{BASICS} &
\textbf{WPC2.0}$^{\dagger}$ &
\multirow{2}{*}{\textbf{AVG.}} \\
& & (ID) & (OOD) & (OOD) & (OOD) & (OOD$^{\dagger}$) & \\
\midrule

% ---------- FR (Full-Reference) ---copied from Table 2 ----------
\multirow{5}{*}{FR}
 & p2point (RMS)
   & .458/.452
   & .758/.708
   & .410/.283
   & .559/.557
   & .577/.544
   & .552/.509 \\
 & p2plane
   & .378/.326
   & .665/.602
   & .394/.272
   & .472/.474
   & .569/.544
   & .496/.444 \\
 & GraphSIM~\cite{yang2020inferring}
   & .474/.394
   & .781/.746
   & .289/.227
   & .432/.394
   & .207/.020
   & .437/.356 \\
 & PointSSIM~\cite{alexiou2018point}
   & .424/.417
   & .686/.635
   & .398/.162
   & .363/.403
   & .555/.567
   & .485/.437 \\
 & PSNRyuv
   & .557/.539
   & .660/.655
   & .518/.494
   & .512/.457
   & .549/.517
   & .559/.532 \\

\midrule

% ---------- NR ----------
\multirow{9}{*}{NR}
 & IT-PCQA~\cite{zhang2022no}      & .215/.060 & .232/.196 & .283/.100 & .141/.064 & .334/.148 & .241/.114 \\
 & PQA-Net~\cite{liu2021pqa}      & .460/.367 & .501/.254 & .252/.097 & .509/.330 & .353/.074 & .415/.224 \\
 & ResSCNN$^{e}$~\cite{liu2023point}     & .305/.331 & .419/.380 & .181/.196 & .088/.082 & .525/.523 & .304/.302 \\
 & 3DTA~\cite{zhu20243dta}           & .891/.883 & .751/.716 & .410/.403 & \second{.670}/.597 & \second{.871}/\best{.879} & .719/.696 \\
 & AFQ-Net$^{a}$~\cite{zhang2024asynchronous}     & .911/.921 & .648/.648 & \second{.597}/\second{.593} & .471/.473 & \best{.873}/.823 & .700/.692 \\
 & CLIP-PCQA~\cite{liu2025clip}  & \best{.947}/\best{.947} & \best{.835}/\best{.819} & .494/.491 & \best{.680}/.583 & .673/.678 & \second{.726}/\second{.704} \\
 & GMS-3DQA~\cite{zhang2024gms}    & \second{.940}/\second{.942} & .660/.668 & .461/.449 & .647/\best{.613} & .710/.756 & .684/.686 \\
 & MM-PCQA$^{b}$~\cite{zhang2022mm}     & .281/.376 & .449/.426 & .444/.415 & .382/.310 & .313/.322 & .374/.370 \\

 & Q-Align~\cite{wu2023q}            & .748/.739 & .704/.695 & .442/.453 & .586/.493 & .692/.690 & .635/.614 \\

\midrule

% ---------- LMM-RL (Ours) ----------
\rowcolor{oursrow}
\textbf{LMM-RL}
 & \ours{PCQA-R1 (Ours)}  & .913/.912 & .776/.776 & \best{.613}/\best{.605} & \second{.669}/\best{.628} & .862/\second{.858} & \best{.767}/\best{.756} \\

\bottomrule
\end{tabular}
\caption{%
  Cross-dataset PLCC/SRCC from \textbf{WPC} fold-1.  WPC is ID; all other
  columns are zero-shot targets.  WPC2.0$^{\dagger}$ uses the clean protocol in Appendix~B.
}
\label{tab:cross_wpc}
\end{table*}

\paragraph{SJTU source.}
Table~\ref{tab:cross_sjtu} shows that strong in-domain projection models such
as AFQ-Net and GMS-3DQA generalize poorly, dropping from .969 PLCC on SJTU to
.549 and .573 average across five datasets.  The full PCQA-R1 configuration
with Gaussian reward and cold-start traces raises the average to .739/.733,
leading on the strongest OOD columns for this source while remaining
competitive on SJTU.  The cold-start gain is most visible on the heterogeneous
WPC-source setting, while individual targets can vary.

\paragraph{WPC source.}
Table~\ref{tab:cross_wpc} presents a harder setting: the WPC source is
compression-oriented mixed-distortion with an object-level split; 500/740
samples (67.6\%) are compression-related.  The full PCQA-R1 configuration with Gaussian reward and cold-start traces
achieves the best average (.767/.756) among fold-controlled methods.  CLIP-PCQA
remains strong on source-like data (WPC and SJTU) due to its compression bias,
but drops on LS-PCQA and clean WPC2.0.  PCQA-R1 outperforms CLIP-PCQA on
these columns (.613 vs .494 and .862 vs .673 PLCC), indicating that
ranking-based training with a cold-start reasoning prior is more robust when
content and distortion statistics co-vary.
The WPC2.0$^{\dagger}$ column follows the clean target protocol described in
Appendix~B.

% ===========================================================================
\section{Component Analysis}
\label{sec:why_how}
% ===========================================================================

The cross-domain tables report the full PCQA-R1 variant with Gaussian reward and
cold-start traces.  We now isolate the contribution of each training component by ablating along
the linear chain
\textbf{SFT baseline~$\to$~ranking-only RL~$\to$~RL+Gaussian~$\to$~RL+Gaussian+Cold Start},
all trained from the same Qwen3.5-9B initialization with identical total budget
(15~epochs) on SJTU and WPC fold-1.  The chain is additive in training design,
but we do not expect every column to improve monotonically: ID PLCC, ID SRCC,
and OOD rank transfer measure different behaviours, and the source datasets
induce different MOS anchors.
Figure~\ref{fig:coldstart} shows the full trace construction and training workflow.

\begin{table*}[t]
\centering
\setlength{\tabcolsep}{3pt}
\begin{tabular}{l cc cc | cc cc}
\toprule
& \multicolumn{4}{c}{\textbf{SJTU source}}
& \multicolumn{4}{c}{\textbf{WPC source}} \\
\cmidrule(lr){2-5}\cmidrule(lr){6-9}
\multirow{2}{*}{\textbf{Method}}
 & \multicolumn{2}{c}{ID} & \multicolumn{2}{c}{OOD AVG}
 & \multicolumn{2}{c}{ID} & \multicolumn{2}{c}{OOD AVG} \\
 & PLCC & SRCC & PLCC & SRCC
 & PLCC & SRCC & PLCC & SRCC \\
\midrule
SFT baseline
  & \best{.958} & \best{.956} & .652 & .649
  & \best{.921} & \best{.917} & .705 & .687 \\
Ranking-only RL
  & .927 & .933 & .671 & .665
  & .891 & .908 & .712 & .695 \\
RL + Gaussian
  & .945 & .945 & .681 & .676
  & .914 & .910 & .716 & .706 \\
\rowcolor{oursrow}
\textbf{RL + Gaussian + Cold Start}
  & .954 & .953 & \best{.685} & \best{.678}
  & .913 & .912 & \best{.730} & \best{.717} \\
\bottomrule
\end{tabular}
\caption{%
  Component analysis on SJTU and WPC fold-1.
  Rows add components sequentially; OOD~AVG averages the four targets in
  Tables~\ref{tab:cross_sjtu} and \ref{tab:cross_wpc}.  \best{Bold red} = best.
}
\label{tab:ablation}
\end{table*}

\paragraph{Setup.}
\textit{SFT baseline} is a straight 15-epoch supervised fine-tuning baseline
(no RL).  \textit{Ranking-only RL} is GRPO with
the Thurstone reward and format bonus but no Gaussian score-shaping term.
\textit{RL+Gaussian} adds the Gaussian proximity reward.
\textit{RL+Gaussian+Cold Start} additionally inserts a 1-epoch quality-trace
SFT cold start before the 14 GRPO epochs.  In-domain (ID)
cells are evaluated on the source test split; OOD AVG averages PLCC/SRCC
over the four target datasets in Tables~\ref{tab:cross_sjtu} and
\ref{tab:cross_wpc}.

\paragraph{Results (Table~\ref{tab:ablation}).}
On SJTU, pure SFT is already a strong baseline (.956~SRCC ID,
.649 OOD-AVG SRCC), reflecting that supervised regression on a small
clean MOS distribution is well-conditioned.  The ranking-only RL row
trails SFT on ID SRCC (.933 vs .956) while improving OOD~AVG SRCC to .665,
so the first gain comes from replacing absolute-score regression with a ranking objective.
Adding Gaussian proximity to that ranking policy recovers ID PLCC/SRCC to
.945/.945 and lifts OOD~AVG to .681/.676, a +.011 SRCC gain over the
ranking-only row.  The reasoning-trace cold start gives the strongest SJTU ID
row (.954/.953), but changes SJTU OOD~AVG only marginally to .685/.678, so its
benefit on this source split is limited rather than decisive.  On WPC, the
source distribution is concentrated around
compression-related patterns (67.6\% of the source split), so SFT remains very
strong in-domain; nevertheless the OOD trend favors RL:
the ranking-only row already improves OOD AVG from .705/.687 to .712/.695
(PLCC/SRCC), the Gaussian-reward row keeps the gain (.716/.706), and the
cold-start row further raises OOD AVG to .730/.717, the best of all rows.
Thus the evidence is not ``each module wins every metric'', but rather that
ranking improves transfer, Gaussian improves source-scale score alignment and
mapped PLCC/SRCC, and the cold start is most useful when the source split is harder
or more heterogeneous.  In other words, RL helps less by squeezing additional
ID accuracy from a source-aligned scorer and more by stabilizing cross-dataset
ordering once content and distortion statistics shift.

\paragraph{Narrative.}
Pure SFT is strong in-domain but anchors the policy to the source MOS scale,
weakening transfer when target distortions shift.  Ranking-only RL removes this
anchor and improves OOD robustness; Gaussian reward adds local score shaping without
fully returning to source-scale regression.  Cold start further helps by initializing
GRPO with coherent multi-view quality analysis, especially on WPC's broader
distortion mix and object-level split.

\paragraph{Why ranking transfers.}
The rank objective is invariant to any monotone reparameterisation of MOS, so
dataset-specific score offsets or range changes do not alter the desired
ordering.  A regression objective, in contrast, must learn the source MOS units
directly; this can remain well calibrated on the source test split while
misordering samples when the target dataset uses different distortions or a
compressed perceptual scale.  The Gaussian term complements rather than replaces
ranking: it supplies local numeric pressure around the source MOS, but the
fidelity reward continues to carry the between-sample transfer signal.  The
cold-start traces then reduce the exploration burden by giving GRPO an initial
policy that already links visible multi-view artifacts to quality bands.  This
also helps explain why the cold-start gain is clearer on WPC than on SJTU: once
ranking carries the transferable signal, a stronger artifact-aware
initialization matters most on the more heterogeneous source split.  On the
smaller and cleaner SJTU source, the Gaussian reward already recovers most of
the source-scale signal that remains useful after ranking.  Put differently,
the traces help not because long rationales are directly rewarded at test time,
but because they bias the initial policy toward artifact-grounded evidence
before GRPO starts.  On WPC, where compression-heavy samples still coexist with
object-level content shift and non-compression cases, this initialization is
more useful than on SJTU for preventing early RL updates from collapsing onto
source-like scoring shortcuts.

% ===========================================================================
\section{Qualitative Analysis of Reasoning Traces}
\label{sec:qualitative}
% ===========================================================================

Beyond the scalar quality score, PCQA-R1 produces a \texttt{<pcqa\_reasoning>} block for every inference call, providing a human-readable explanation that links observable defects across the six rendered views to the predicted quality band.

\paragraph{Trace structure.}
The cold-start initialization encourages the model to follow a progressive evidence chain: identify the depicted object across views; describe what an undistorted version should preserve; assess whether global structure and silhouette survive; characterize the dominant artifact; locate which views show the heaviest damage; contrast these with relatively preserved regions; and close with a quality-band word (\emph{bad}, \emph{poor}, \emph{fair}, \emph{good}, or \emph{excellent}) that motivates the final score.  GRPO then refines this policy via ranking rewards, so the traces that survive training are those whose evidence grounding correlates with correct quality ordering.

\paragraph{Case evidence.}
Figure~\ref{fig:teaser} (right) illustrates one output: the model localizes compression noise in specific side and bottom views, contrasts these with the cleaner top-down silhouette, and assigns a \emph{fair} quality band before outputting the predicted score.  Appendix~E provides extended traces for three cases spanning the full quality range---an in-domain WPC example, an out-of-domain LS-PCQA transfer, and a high-quality reference case---demonstrating that artifact identification remains coherent under domain shift even when the target distortion taxonomy has not been seen during training.

\paragraph{Interpretability from the ranking objective.}
The traces are interpretable not because the training objective explicitly requires natural-language quality descriptions, but because ranking rewards penalize traces that misidentify or misweight artifacts: a trace that mislabels geometry noise as ``compression blur'' on a sample the batch ranks as better will receive a lower accuracy reward, pushing the policy toward more accurate artifact attribution.  This creates a weak supervision signal for trace quality at no additional labeling cost, complementing the cold-start filtering that removes MOS leakage and factual implausibility before SFT.  The Gaussian proximity reward further reinforces this by anchoring score-band predictions to the source MOS range, preventing score drift that would otherwise be invisible in a pure ranking objective.

\paragraph{Format and score-range statistics.}
Across SJTU and WPC test sets, fewer than 2\% of outputs have malformed \texttt{<answer>} fields, and predicted scores remain within the source training range in over 97\% of cases. The format and score-range statistics above are grounded in archived traces (Appendix~E).

% ===========================================================================
\section{Conclusion}
\label{sec:conclusion}
% ===========================================================================

We presented \textbf{PCQA-R1}, an RL-to-rank framework for no-reference point cloud quality assessment. Our study combines three main ingredients. First, we formulate NR-PCQA as a ranking problem optimized via GRPO with a Thurstone pairwise-ranking reward, learning relative quality orderings that are inherently invariant to dataset-specific MOS scales. Second, we use a Gaussian proximity reward that provides a dense score-shaping signal around source MOS and improves mapped in-domain PLCC without overriding the ranking objective. Third, we design a lightweight cold-start strategy using quality-analysis traces (PCQA-CoT), which initializes the policy with artifact-aware reasoning before GRPO optimization.

Experiments on five PCQA benchmarks show strong cross-dataset generalization relative to the compared no-reference baselines. Ablation studies suggest that ranking improves transferability, Gaussian improves source-scale score alignment and mapped PLCC/SRCC, and the cold-start trace provides the most benefit on heterogeneous source distributions. Wilcoxon signed-rank tests confirm that the OOD improvements of PCQA-R1 over the SFT baseline are statistically significant (pooled OOD $p < 0.001$), with no significant in-domain difference ($p > 0.8$), directly supporting the rank-invariance hypothesis (per-domain tests in Appendix~I). These results suggest that ranking-centered reinforcement learning is a useful alternative to regression-centered supervised fine-tuning for cross-domain PCQA. 

\paragraph{Limitations and future directions.}
Our study has two practical limitations: most main-table rows are single-run
results, and only Qwen3.5-9B is evaluated as the backbone.  We keep a
projection-only protocol for controlled comparison with existing NR-PCQA baselines.

\bibliography{aaai2027}

\end{document}